\documentclass[11pt]{article}
\usepackage[preprint]{acl}
\usepackage{times}
\usepackage{latexsym}
\usepackage[T1]{fontenc}
\usepackage[utf8]{inputenc}
\usepackage{microtype}
\usepackage{inconsolata}
\usepackage{graphicx}
\usepackage{amsmath}
\usepackage{amssymb}
\usepackage{booktabs}
\usepackage{array}
\usepackage{tabularx}
\usepackage{multirow}
\usepackage{xspace}
\usepackage{listings}
\usepackage{algorithm}
\usepackage{algpseudocode}
\usepackage{fontawesome5}

\definecolor{tblHead}{HTML}{E8EEF5}
\definecolor{slotEntity}{HTML}{DCEBFA}
\definecolor{slotType}{HTML}{E5DDF7}
\definecolor{slotProp}{HTML}{FBE5C8}
\definecolor{slotValue}{HTML}{DDF2E1}
\definecolor{slotRel}{HTML}{F7DDE5}
\definecolor{slotState}{HTML}{D8EDEF}
\definecolor{slotCond}{HTML}{F5E8B8}

\newcommand{\slot}[2]{%
  \begingroup
  \setlength{\fboxsep}{2pt}%
  \colorbox{#1}{\strut #2}%
  \endgroup
}

\newcommand{\slotsep}{\;\textbar\;}
\newcommand{\slotarrow}{$\rightarrow$}
\newcommand{\softrule}{\addlinespace[2pt]\midrule}

\newcommand{\method}{\textsc{Resin}\xspace}

\definecolor{promptBlue}{HTML}{DCEBFA}
\definecolor{promptGreen}{HTML}{DDF2E1}
\definecolor{promptBg}{HTML}{F4F7FB}
\lstdefinestyle{ensicode}{
  basicstyle=\ttfamily\footnotesize,
  columns=fullflexible,
  keepspaces=true,
  breaklines=true,
  breakatwhitespace=true,
  frame=single,
  framesep=5pt,
  rulecolor=\color{promptBlue},
  backgroundcolor=\color{promptBg},
  xleftmargin=2pt,
  xrightmargin=2pt,
  aboveskip=3pt,
  belowskip=8pt,
  showstringspaces=false
}
\lstdefinestyle{ensialg}{
  style=ensicode,
  basicstyle=\ttfamily\scriptsize,
  numberstyle=\tiny,
  numbersep=5pt,
  aboveskip=2pt,
  belowskip=5pt
}
\lstdefinestyle{promptcode}{
  style=ensicode,
  breakatwhitespace=false,
  rulecolor=\color{promptGreen!75!black},
  backgroundcolor=\color{promptBg},
  xleftmargin=0pt,
  xrightmargin=0pt,
  aboveskip=0pt,
  belowskip=9pt
}

\newcommand{\prompttitle}[1]{%
  \noindent\colorbox{promptGreen}{%
    \parbox{\dimexpr\linewidth-2\fboxsep\relax}{\strut\textbf{#1}}}}
\title{RESIN: Multi-Chain Reasoning with Entity-Structured Indexing}
\author{
  Xuanyu Meng \quad Jash Rajesh Parekh \quad Jiawei Han \\
  University of Illinois Urbana-Champaign \\
  \texttt{\{xuanyum2,jashrp2,hanj\}@illinois.edu}
}

\newcommand{\nop}[1]{}

\begin{document}
\maketitle

\begin{abstract}
\nop{
Multi-chain questions are difficult not only because their evidence is dispersed, but also because different parts of an answer may require distinct, interacting chains of support. Flat retrieval can flood a model with loosely related text, whereas early selection of a single reasoning path can discard evidence that becomes useful only after other facts are considered. We introduce \method (Reasoning with Entity Structured Index), which uses structure to \emph{locate} evidence while keeping the original source as the basis for reasoning. \method decomposes a question into focused one-hop needs, retrieves a small set of source-grounded units for each, and places only this relevant evidence in a compact reasoning buffer. The buffer preserves alternative supports until final composition, making intermediate decisions explicit without forcing a brittle path. On the full HolisQA benchmark, \method reaches 82.87 context recall and 76.52 answer relevancy, with consistent gains across five domains. Ablations show that both typed evidence diversity and hop-wise decomposition are essential, supporting a simple conclusion: precise evidence organization and delayed commitment make multi-chain reasoning more effective and auditable.}  
Multi-chain reasoning is a challenging task not only because the multi-chain evidence is dispersed, but also because different parts of reasoning may require distinct, interacting chains of support. Flat retrieval can flood a model with loosely related text, whereas early selection of a single reasoning path can discard evidence that becomes useful only after other facts are considered. We introduce \method (Reasoning with Entity Structured Indexing), which uses structure to \emph{locate evidence} while \textit{keeping the original source as the basis for reasoning}. \method decomposes a question into focused one-hop needs, retrieves a small set of source-grounded units for each hop, and places only this relevant evidence in a compact reasoning buffer. The buffer preserves alternative supports until final composition, making intermediate decisions explicit without forcing a brittle path. On the full HolisQA benchmark, \method 
achieves 82.92 context recall and 76.70 answer relevancy over all questions, with higher reported CR and AR across five domains and higher F1 in four domains than the published baselines.
Ablations show that both typed evidence diversity and hop-wise
decomposition are essential, supporting a clear conclusion:
\textit{Precise evidence organization and delayed commitment make
multi-chain reasoning effective and auditable}.
Our code is available at
\href{https://github.com/RamonMeng/RESIN}{%
  \faGithub\,\texttt{github.com/RamonMeng/RESIN}}.
\end{abstract}

\begin{figure*}[t!]
    \centering
    \includegraphics[width=\textwidth]{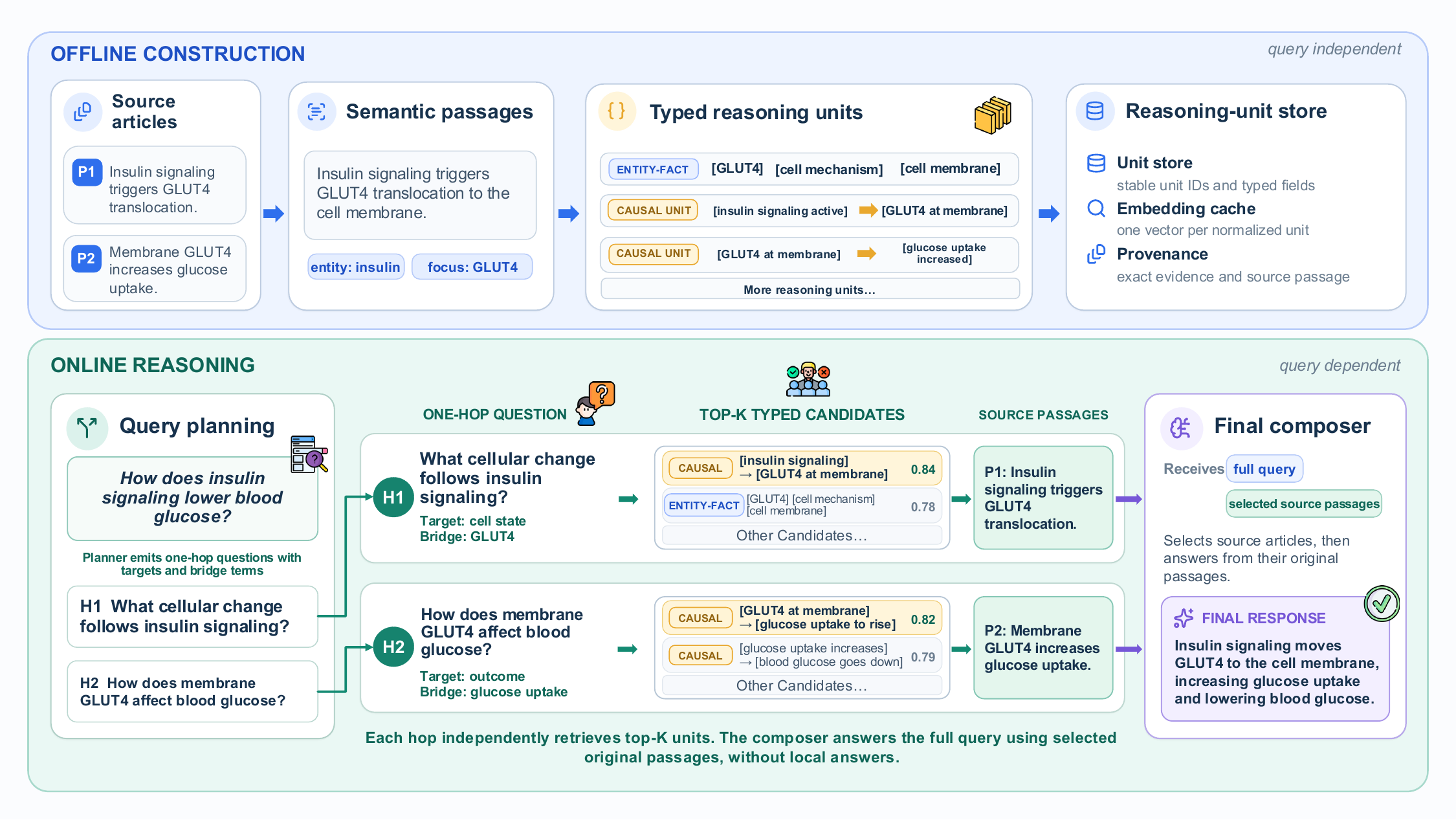}
    \caption{\method at a glance. Offline, entity-structured records serve as addresses into source-grounded evidence. Online, a dependency-aware plan creates focused one-hop needs; each retrieves a small candidate set, and the resulting evidence buffer remains available until final composition.}
    \label{fig:architecture}
\end{figure*}

\section{Introduction}
\nop{ 
Multi-chain reasoning requires a model to assemble several evidence chains whose roles are not interchangeable. One chain may identify an entity, another may explain a state transition, and a third may establish an independent mechanism needed by the same answer. These chains can overlap, diverge, or depend on a bridge entity. The challenge is therefore not simply to find relevant text, but to assign evidence to the right inferential role while preserving entity identity, scope, conditions, and direction.
}

Multi-chain reasoning requires a model to assemble several kinds of evidence chains whose roles are not interchangeable. One kind may identify an entity; another may catch a property or a state transition; and the third may conduct reasoning by following certain rules. The challenge is therefore not simply to find relevant text, but to assign evidence to the right inferential role while preserving entity identity, scope, condition, and direction.

\nop{ 
This distinction matters because textual relevance is not the same as reasoning sufficiency. A passage can share many words with the question yet contribute no missing link, while a less obvious passage may supply the bridge that makes two claims composable. Conversely, two individually plausible statements may refer to different entities, populations, or experimental conditions and therefore cannot safely form one chain. A system that ranks evidence only against the complete question has no explicit way to represent these roles. A system that commits to the first locally plausible chain risks the opposite error: it may explain one part of the answer while losing a parallel chain needed for another.
}

Such distinct mechanisms make multi-chain reasoning a rich but challenging task. A passage that share many words with the question may not eventually contribute to the reasoning path, whereas a less obvious passage may later supply the bridge for effective multi-chain reasoning.  A system that ranks evidence only against the complete question cannot handle such complexity. Thus a system that commits to the first locally plausible chain risks the opposite error: it may explain one part of the answer while losing a parallel chain needed for another.

\nop{ 
Existing evidence-access strategies tend to fail at one of two extremes. A single broad search compresses all information needs into one query, which can blur their roles, introduce distracting context, and impose the same granularity on every step \citep{liu2024lost,chen2024dense,jiang2024longrag}. Repeated search exposes intermediate information needs \citep{trivedi2023ircot,jiang2023flare,asai2024selfrag}, but untyped passages still leave entity and state bindings implicit. Graph-based systems make relations explicit \citep{edge2024graphrag,gutierrez2025hipporag2,li2024structrag,guo2024lightrag}, yet extraction and edge ranking are uncertain: pruning to one node, edge, or path can remove correct support before all chains are reconciled. This is especially risky in scientific text, where a relation may change with the condition, population, or state \citep{parekh2026condition}.
}

Existing evidence-access strategies tend to encounter difficulties at handling such complexity. A single broad search compresses all information needs into one query, which can blur their roles, introduce distracting context, and impose the same granularity on every step \citep{liu2024lost,chen2024dense,jiang2024longrag}. Repeated search exposes intermediate information needs \citep{trivedi2023ircot,jiang2023flare,asai2024selfrag}, but untyped passages still leave entity and state bindings implicit. Graph-based systems make relations explicit \citep{edge2024graphrag,gutierrez2025hipporag2,li2024structrag,guo2024lightrag}, yet extraction and edge ranking are uncertain: pruning to one node, edge, or path can remove correct support before all chains are reconciled. This is especially risky in scientific text, where a relation may change with the condition, population, or state \citep{parekh2026condition}.

The underlying mismatch is between the object that can be indexed in advance and the object required to answer a particular question. Evidence records should be reusable and query independent, whereas a valid chain is query specific: the same statement may be central for one question, a bridge for another, and irrelevant for a third. Precomputing every possible chain is neither tractable nor robust, because the number of combinations grows quickly and every extracted edge becomes an early commitment. Deferring all organization until generation avoids that early commitment but returns the model to a large, weakly structured context. What is needed is an intermediate representation that is fine grained enough to locate precise evidence, yet flexible enough to assemble different chains on demand.

We propose a different division of labor. \emph{Structure should tell the system where to look}; whereas \textit{source-grounded evidence should tell it what to believe}.  \method organizes evidence into typed, provenance-linked units that act as precise addresses. It then builds a small \emph{reasoning buffer} for the current question rather than exposing the model to the full text or index. The buffer assigns a candidate set to each planned hop and retains alternatives until final composition. Thus, structure improves access without becoming a brittle, prematurely fixed reasoning path.

Figure~\ref{fig:architecture} makes this separation concrete. The upper half is query independent: source passages are organized around their central entities, converted into compact reasoning units, and linked back to exact evidence. The lower half begins with the question: a planner expresses it as focused one-hop needs, retrieval fills a bounded candidate set for each need, and local answers expose what each set supports. The final composer sees the original question and the complete buffer. Dependency metadata records how the subquestions fit together, while the evidence itself remains available for global reconciliation.

This architecture offers three practical advantages. First, it bounds the model's working context by the number of planned needs and the per-hop candidate budget, rather than by corpus size. Second, it makes failures inspectable: one can distinguish a missing indexed unit, a retrieval miss, an incorrect local interpretation, and a composition error. Third, it keeps competing evidence recoverable until the system has seen all parts of the question. These properties motivate our central hypothesis: multi-chain reasoning improves when evidence is localized early but selected late. The experiments test both sides of this claim by varying the unit representation, the decomposition strategy, and the number of candidates retained per hop.

Our contributions are:
\begin{itemize}
    \item an entity-structured index that localizes composable, typed evidence while preserving exact source provenance;
    \item a hop-wise reasoning buffer that retrieves focused candidate sets and delays hard evidence selection until the complete question can be considered; and
    \item evaluation on HolisQA and two general-domain QA benchmarks, with ablations showing when decomposition, evidence diversity, and retrieval budget matter.
\end{itemize}

\section{Related Work}

\paragraph{From passage retrieval to focused information needs.}
RAG combines parametric generation with external retrieval \citep{lewis2020rag}, and dense passage retrieval provides a strong semantic interface for open-domain QA \citep{karpukhin2020dpr}. Work on Dense X Retrieval, LongRAG, and RAPTOR shows that retrieval granularity and hierarchy materially affect what evidence reaches the model \citep{chen2024dense,jiang2024longrag,sarthi2024raptor}. These methods improve access to long corpora, but a retrieved passage collection still asks the generator to discover which part serves which reasoning step. \method instead starts from explicit information needs and gives each one a small, source-grounded evidence set.

\paragraph{Iterative retrieval and multi-step reasoning.}
Chain-of-thought and ReAct make intermediate decisions visible \citep{wei2022cot,yao2023react}. IRCoT alternates retrieval with reasoning, FLARE retrieves during generation, and Self-RAG learns when to retrieve and critique \citep{trivedi2023ircot,jiang2023flare,asai2024selfrag}. Their central insight---reasoning should guide evidence access---also motivates \method. Our distinction is representational: planned hops search a typed index, and all retrieved alternatives remain in a bounded buffer for final reconciliation instead of being reduced immediately to a single intermediate statement.

\paragraph{Structured, entity-centered, and causal retrieval.}
GraphRAG, HippoRAG2, StructRAG, and LightRAG organize entities and relations to support global, associative, or structure-aware retrieval \citep{edge2024graphrag,gutierrez2025hipporag2,li2024structrag,guo2024lightrag}. SARG searches explicit relational chains \citep{parekh2025sarg}, while CausalRAG2 builds hierarchical causal graphs and applies top-down pruning for HolisQA \citep{wang2026causalrag2}. Such structure is valuable, but a selected path can conflate indexing with reasoning: an extraction or pruning error may remove evidence before the answer is composed. Entity structure has also proved useful in other settings: EnSI-RAG uses it to connect properties and relations to source passages \citep{meng2026ensirag}, EnSIMem uses it to organize verifiable long-term agent memories \citep{meng2026ensimem}, and open-schema discovery extracts entity--attribute--value structure without a fixed ontology \citep{xu2026zoes}. These applications illustrate the general utility of entities as stable access keys, but address different tasks. \method targets multi-chain reasoning by using typed structure as an \emph{address} into evidence and retaining a small buffer of source-linked alternatives, rather than treating the structure as an irrevocable answer path.

\section{RESIN}

\subsection{Design Goal and Running Example}

Given corpus $D$ and question $q$, \method uses entity structure to \emph{find the relevant original text}, then reasons directly over that text. Offline, it builds searchable, source-linked records. Online, it plans focused searches, retrieves records for each need, recovers their source passages, and composes an answer from a small selected evidence buffer (Figure~\ref{fig:architecture}).

Consider an illustrative commute: \emph{Should I take the bus or metro to reach work within 40 minutes and spend at most \$10?} Suppose route descriptions list the bus at \$4, with a 10-minute walk and a 35-minute ride; the metro costs \$8, with a 5-minute walk and a 25-minute ride. Assuming no waiting, the bus takes 45 minutes and the metro 30. Only the metro satisfies both constraints. Answering requires separate evidence chains for the two routes, keeping fares and travel times attached to the correct option.

\subsection{Design Rationale}

\paragraph{Give each answer component a search.}
The planner asks separately for the bus's fare and journey times and for the metro's corresponding values. Each need gets its own candidate set, so many matches about one route cannot consume the entire retrieval budget. The hops identify what must be found; they do not decide which route to recommend.

\paragraph{Keep facts attached to their meaning.}
A record binds a value to its entity, property, and qualifiers: \emph{the bus ride takes 35 minutes}, not simply \emph{35}. This distinguishes ride time from total journey time and bus facts from metro facts. Its source pointer recovers the wording needed to check those distinctions.

\paragraph{Make the decision from original evidence.}
The final composer receives the original question and selected source passages, not generated hop answers. It calculates each journey's total time and checks both constraints together. An early conclusion such as ``the bus is cheaper'' therefore need not determine the recommendation before arrival time is considered.

\subsection{Source-Linked Evidence Organization}

Offline processing groups contiguous sentences around an entity and evidence role, preserves their text and sentence IDs, and extracts minimal evidence-backed units. The bus's ride-time statement becomes a searchable record linked to its route passage; unrelated service announcements need not enter that record. The record is an access handle, while the original passage remains the evidence.

Table~\ref{tab:unit-schema} groups fields by function: claim identity, composition constraints, qualifiers, and provenance. Optional fields require source support.
\begin{table}[t]
\centering
\footnotesize
\setlength{\tabcolsep}{3.5pt}
\renewcommand{\arraystretch}{1.18}
\begin{tabularx}{\columnwidth}{@{}p{0.19\columnwidth}p{0.29\columnwidth}X@{}}
\toprule
\textbf{Field group} & \textbf{Stored fields} & \textbf{Purpose and example} \\
\midrule
Identity & kind; stable unit ID & Distinguishes a fact from a rule or directed transition. \\
Core claim & subject; entity type; predicate; object & Binds ``bus fare is \$4'' to the correct route and property. \\
Logical structure & conditions and conclusion; source and target states & Preserves conditions and direction, e.g., congestion increasing ride time. \\
Qualifiers & modality; polarity; scope; confidence & Keeps a weekday travel-time estimate distinct from an unrestricted claim. \\
Grounding & evidence quote; passage and sentence IDs & Links every normalized record back to the exact source span. \\
\bottomrule
\end{tabularx}
\caption{Functional schema of a reasoning unit. Optional fields are populated only when supported by the source.}
\label{tab:unit-schema}
\end{table}

Table~\ref{tab:unit-kinds} distinguishes six kinds of support. A route's fare is an entity fact; a source-stated effect of congestion on ride time is a directed transition. Rules retain their conditions, and generic type properties remain separate from entity membership. Only source-supported fields are extracted: journey totals are computed at answer time, not inserted as facts during indexing.
\begin{table*}[t]
\centering
\small
\setlength{\tabcolsep}{6pt}
\setlength{\aboverulesep}{0pt}
\setlength{\belowrulesep}{0pt}
\caption{Reasoning-unit kinds and canonical representations. Colors mark entity, type, property, value, relation, state, and condition slots.}
\label{tab:unit-kinds}
\renewcommand{\arraystretch}{1.5}
\begin{tabularx}{\textwidth}{@{}
    >{\raggedright\arraybackslash}p{0.15\textwidth}
    >{\raggedright\arraybackslash}p{0.38\textwidth}
    >{\raggedright\arraybackslash}X@{}}
\toprule
\textbf{Unit kind} & \textbf{Representation} & \textbf{Why it is useful} \\
\midrule
\textsc{entity-fact}
  & \slot{slotEntity}{entity}\slotsep\slot{slotProp}{property}\slotsep\slot{slotValue}{value}
  & Retrieves an attribute stated for one named entity. \\
\softrule
\textsc{type-property}
  & \slot{slotType}{type}\slotsep\slot{slotProp}{property}\slotsep\slot{slotValue}{value}
  & Stores a generic property that may support type inheritance. \\
\softrule
\textsc{entity-type}
  & \slot{slotEntity}{entity}\slotsep\slot{slotRel}{is-a}\slotsep\slot{slotType}{type}
  & Supplies explicit membership evidence before a type property is applied. \\
\softrule
\textsc{relation}
  & \slot{slotEntity}{entity or type}\slotsep\slot{slotRel}{relation}\slotsep\slot{slotEntity}{entity or concept}
  & Preserves a typed link that is neither causal nor conditional. \\
\softrule
\textsc{causal-edge}
  & \slot{slotState}{source state}\slotarrow\slot{slotState}{target state}
  & Keeps causal direction and state change explicit. \\
\softrule
\textsc{rule}
  & \slot{slotCond}{conditions}\slotarrow\slot{slotCond}{conclusion}
  & Prevents a conclusion from being used when its antecedents are absent. \\
\bottomrule
\end{tabularx}
\end{table*}

The fields guide the LLM, not a symbolic proof checker: they distinguish evidence roles and expose direction, binding, and qualification for inspection.

\subsection{The Index as an Evidence Address}

The searchable store is
\begin{equation}
\mathcal{U}=\{(\operatorname{id}(u),x_u,\pi_u):u\in U\},
\label{eq:unit-store}
\end{equation}
where $x_u$ is normalized retrieval text and $\pi_u$ links to the evidence quote, passage, and source sentences. Auxiliary entity, type, property, state, and relation views provide lookup handles. A sentence encoder caches embeddings of $x_u$; the main online backend uses dense unit retrieval, not graph traversal.

In Equation~\ref{eq:unit-store}, $x_u$ makes ``bus ride time'' searchable, and $\pi_u$ leads back to the route passage. This separates finding a relevant fact from deciding how to use it; bus and metro records need no pre-existing connection in the index.

\subsection{Planning and Need-Specific Retrieval}

A planner maps $q$ to an ordered plan $Q=(q_1,\ldots,q_H)$. Each hop specifies an evidence need, target and bridge terms, a reasoning type, and optional dependencies. Here, $q_1$ asks for the bus's fare, walk, and ride times; $q_2$ asks for the metro's. All hops are planned upfront and retrieve independently: the second does \emph{not} use an answer generated for the first. Failed planning falls back to the original question as one hop.

For retrieval text $\tilde q_h$ formed from the subquestion, target, and bridge terms, we select
\begin{equation}
C_h=\operatorname{TopK}_{u\in U}\cos\!\left(E(\tilde q_h),E(x_u)\right).
\label{eq:hop-retrieval}
\end{equation}
Equation~\ref{eq:hop-retrieval} retrieves bus records into $C_1$ and metro records into $C_2$, with a separate top-$k$ budget for each need. Retrieval returns evidence candidates, not route recommendations. The main setting permits at most four hops and $k=10$ units per hop.

Multiple hops also increase total capacity; the decomposition ablation is not budget-matched. Upfront planning can miss a bridge entity not expressible from the question.

\subsection{Buffer Construction and Final Composition}

Provenance maps the retrieved units to original passages. Let $P(q)$ be these passages, deduplicated and packed by alternating across hops at each candidate rank within a 16,000-character budget. The final evidence buffer is
\begin{equation}
\mathcal{B}(q)=\operatorname{SelectSources}\!\left(q,P(q)\right),
\label{eq:buffer}
\end{equation}
where the selector uses the question, article titles, and source snippets to choose up to two article groups and retain their packed passages. A failed selection falls back to question-term ranking of up to three groups. Selection uses neither gold answers nor generated hop answers.

The composer receives only $q$ and $\mathcal{B}(q)$, not the query plan or intermediate answers. From both route passages, it computes $10+35=45$ minutes for the bus and $5+25=30$ for the metro. Both fares are within budget, but only the metro meets the deadline. The hops found the inputs; final composition makes the comparison. Auxiliary hop-solver outputs are still recorded by the implementation but are unused in this generation path. Evidence lost during extraction, retrieval, or source selection remains a limitation.

\section{Experiments}

\begin{table*}[t]
\centering
\footnotesize
\setlength{\tabcolsep}{2.3pt}
\caption{HolisQA results across five domains and all questions. F1 measures answer overlap; CR and AR denote context recall and answer relevancy. All scores are percentages. Baseline scores are transcribed from the published CausalRAG2 table; only the \method row is measured in our experiment. Highest and second-highest values are \textbf{bold} and \underline{underlined}; cross-system evaluation caveats are discussed in the text.}
\label{tab:holisqa-main}
\begin{tabular*}{\textwidth}{@{\extracolsep{\fill}}lrrr|rrr|rrr|rrr|rrr@{}}
\toprule
& \multicolumn{3}{c|}{Medicine} & \multicolumn{3}{c|}{Computer Science} &
\multicolumn{3}{c|}{Business} & \multicolumn{3}{c|}{Biology} &
\multicolumn{3}{c}{Psychology} \\
\cmidrule(lr){2-4}\cmidrule(lr){5-7}\cmidrule(lr){8-10}\cmidrule(lr){11-13}\cmidrule(lr){14-16}
Method & F1 & CR & AR & F1 & CR & AR & F1 & CR & AR & F1 & CR & AR & F1 & CR & AR \\
\midrule
\multicolumn{16}{c}{\emph{Naive baselines}} \\
NaiveGeneration & 12.63 & 0.00 & 44.70 & 18.93 & 0.00 & 48.79 & 18.58 & 0.00 & 46.14 & 11.71 & 0.00 & 45.76 & 22.91 & 0.00 & 50.00 \\
BM25            & 17.72 & 52.04 & 50.64 & 24.00 & 39.12 & 52.40 & 28.11 & 37.06 & 55.52 & 19.61 & 43.02 & 52.32 & 30.46 & 33.44 & 56.63 \\
StandardRAG     & 26.87 & 61.08 & 56.24 & 28.87 & 49.44 & 57.10 & 47.57 & 46.79 & 67.42 & 28.31 & 42.69 & 57.58 & 37.19 & 52.21 & 59.85 \\
\midrule
\multicolumn{16}{c}{\emph{Graph-based RAG}} \\
GraphRAG Global & 17.13 & 54.56 & 48.19 & 23.75 & 37.65 & 53.17 & 23.62 & 25.01 & 48.12 & 20.67 & 40.90 & 52.41 & 31.09 & 34.26 & 54.62 \\
GraphRAG Local  & 19.03 & 56.07 & 49.52 & 25.10 & 39.90 & 53.30 & 25.01 & 27.36 & 49.05 & 22.21 & 41.88 & 52.73 & 32.31 & 35.22 & 55.02 \\
LightRAG        & 12.16 & 52.38 & 44.15 & 22.59 & 41.86 & 51.62 & 29.98 & 34.22 & 54.50 & 17.70 & 41.24 & 50.32 & 33.63 & 45.54 & 56.42 \\
\midrule
\multicolumn{16}{c}{\emph{Structural / causal augmented}} \\
HippoRAG2       & 21.12 & 57.50 & 51.08 & 16.94 & 21.05 & 47.29 & 21.10 & 18.34 & 45.83 & 12.60 & 16.85 & 44.56 & 20.10 & 34.13 & 46.77 \\
LeanRAG         & 34.25 & 60.43 & 56.60 & 30.51 & 57.61 & 55.45 & 48.30 & 59.29 & 60.35 & 33.82 & 58.43 & 56.10 & 42.85 & 57.46 & 58.65 \\
CausalRAG       & 31.12 & 58.90 & 58.77 & 30.98 & 54.10 & 57.54 & 45.20 & 44.55 & 66.10 & 33.50 & 51.20 & 58.90 & 42.80 & 55.60 & 61.90 \\
CausalRAG2      & \underline{36.45} & \underline{69.91} & \underline{60.65} & \underline{31.60} & \underline{60.94} & \underline{58.34} &
\textbf{51.51} & \underline{67.34} & \underline{68.76} & \underline{34.80} & \underline{61.97} & \underline{59.99} &
\underline{44.42} & \underline{60.87} & \underline{63.53} \\
\midrule
\method         & \textbf{45.43} & \textbf{80.95} & \textbf{75.43} & \textbf{50.96} & \textbf{86.27} & \textbf{75.09} &
\underline{49.61} & \textbf{84.97} & \textbf{79.05} & \textbf{47.21} & \textbf{84.86} & \textbf{75.86} &
\textbf{46.48} & \textbf{77.95} & \textbf{77.93} \\
\bottomrule
\end{tabular*}
\end{table*}

\begin{table*}[!t]
\centering
\footnotesize
\caption{General-domain results. The \method row uses 500 MS~MARCO queries and 426 QASC queries; baseline F1/CR/AR are transcribed from the published CausalRAG2 full-benchmark table. Highest and second-highest values are \textbf{bold} and \underline{underlined}.}
\label{tab:general-main}
\begin{tabular*}{\textwidth}{@{\extracolsep{\fill}}lrrr|rrr@{}}
\toprule
& \multicolumn{3}{c|}{MS MARCO} & \multicolumn{3}{c}{QASC} \\
\cmidrule(lr){2-4}\cmidrule(lr){5-7}
Method & F1 & CR & AR & F1 & CR & AR \\
\midrule
\multicolumn{7}{c}{\emph{Naive baselines}} \\
NaiveGeneration & 5.28 & 0.00 & 15.06 & 2.69 & 0.00 & 13.74 \\
BM25            & 6.97 & 45.78 & 20.33 & 2.49 & 6.12 & 13.17 \\
StandardRAG     & 14.93 & 48.55 & 31.11 & 2.01 & 5.50 & 13.16 \\
\midrule
\multicolumn{7}{c}{\emph{Graph-based RAG}} \\
GraphRAG Global & 9.41 & 3.65 & 13.08 & 0.68 & 3.38 & 3.56 \\
GraphRAG Local  & 30.87 & 25.71 & 57.76 & 8.30 & 9.54 & 46.59 \\
LightRAG        & 37.70 & 54.22 & 63.54 & 8.20 & 20.40 & 44.35 \\
\midrule
\multicolumn{7}{c}{\emph{Structural / causal augmented}} \\
HippoRAG2       & 23.35 & 45.45 & 55.18 & \underline{14.73} & 4.38 & \underline{49.94} \\
LeanRAG         & 38.02 & 54.01 & 58.49 & 13.19 & 22.80 & 45.51 \\
CausalRAG       & 27.66 & 39.38 & 46.03 & 7.65 & 46.86 & 35.56 \\
CausalRAG2      & \underline{38.40} & \underline{60.48} & \underline{66.02} & 13.35 & \underline{70.80} & 49.40 \\
\midrule
\method & \textbf{38.73} & \textbf{91.65} & \textbf{69.68} & \textbf{37.09} & \textbf{84.51} & \textbf{58.77} \\
\bottomrule
\end{tabular*}
\end{table*}

\subsection{Research Questions}

We ask: (1) Does entity-structured reasoning improve evidence coverage and answer relevance? (2) Does the approach transfer beyond HolisQA? (3) Which parts of the evidence buffer matter? and (4) What costs accompany its more explicit reasoning process?

\subsection{Datasets}

We evaluate on the full HolisQA release \citep{wang2026causalrag2}, 500 queries from MS~MARCO \citep{bajaj2016msmarco}, and 426 queries from QASC \citep{khot2020qasc}. Together, the benchmarks cover scientific synthesis, web-search QA, and two-fact science composition. Appendix~\ref{sec:benchmark-details} gives full benchmark details.

\subsection{Baselines, Metrics, and Setup}

We compare with the published CausalRAG2 suite \citep{wang2026causalrag2}. Baseline accuracy and efficiency entries are transcribed from that paper; only \method is measured in our environment. Appendix~\ref{sec:baseline-details} describes the baselines. On HolisQA, we report answer-overlap F1 alongside RAGAS answer relevancy (AR) and context recall (CR) \citep{es2024ragas}. F1 measures token overlap with the reference answer; AR measures whether an answer addresses the question, whereas CR asks whether retrieved passages cover claims in the reference answer. Appendix~\ref{sec:ragas-metrics} defines AR and CR. All scores are percentages; overall HolisQA AR and CR are query-weighted means across domains.

Our runs use RAGAS~0.4.3 with strictness~3, source-passage contexts, and Gemini-3.5-Flash-Lite as judge. GPT-5-nano performs passage construction, unit extraction, planning, hop solving, and composition; \texttt{all-MiniLM-L6-v2} performs dense retrieval. The main setting allows at most four hops and retrieves $k=10$ units per hop.

\paragraph{Reproduction scope.}
Full-scale CausalRAG2 indexing did not finish within a practical window in our environment. A Biology testing run was completed to verify pipeline execution, not full-dataset performance. We therefore retain published baseline rows. The scores obtained in this test closely match those reported in the CausalRAG2 paper, providing support for the comparability of our evaluation results.

\subsection{Main Results}

\method exceeds the transcribed baselines in reported CR and AR across all five HolisQA domains and in F1 across four domains (Table~\ref{tab:holisqa-main}). Business F1 is 49.61, below CausalRAG2's 51.51; Table~\ref{tab:general-main} reports general-domain CR and AR. CausalRAG2 leads the baseline suite across all five HolisQA domains, making it the main reference for our architectural comparison. We therefore focus the following analysis on what \method adds beyond hierarchical causal retrieval. Evaluator differences and different general-domain query sets remain relevant when interpreting the reported score gaps.

\paragraph{Evidence beyond a causal explanation.}
CausalRAG2 uses hierarchical gates and causal refinement to recover coherent explanations. However, multi-chain questions can also require attributes, experimental observations, and comparative properties that do not form another step in the dominant causal chain. An evidence selection centered on that chain may retain a plausible explanation while omitting support for another requested component. \method makes these heterogeneous claims independently addressable through complementary unit kinds, so their retrieval does not require membership in one connected explanation.

\paragraph{Coverage of distinct information needs.}
A single question-level ranking can allocate much of its evidence budget to the most salient component. Several highly relevant items may then repeat the same mechanism instead of covering different parts of the answer. \method decomposes the question into focused needs and assigns each its own candidate set. This organization gives less prominent components a retrieval opportunity and reduces competition between different evidence roles. The CR drop when decomposition is removed, together with the losses from single-kind retrieval, supports the value of decomposed access and complementary representations.

\paragraph{Joint composition without premature exclusion.}
Evidence can also be lost after retrieval: an early selection may discard a claim whose importance becomes apparent only when another component is considered, or final synthesis may fail to use an available distinguishing fact. \method maps retrieved candidates back to their original passages and gives the final composer the original question and selected source evidence, without the query plan or generated hop answers. The composer can therefore consider explanatory, supporting, and comparative claims together in their original wording rather than inheriting intermediate answer summaries. This changes the reasoning interface from selecting a plausible explanation to assembling support for the question's distinct requirements.

These differences explain why \method is particularly useful when an answer depends on heterogeneous, unevenly salient, or interacting evidence. Appendix~\ref{sec:diagnostic-cases} provides detailed case studies tracing these failure patterns and the corresponding evidence recovered by \method.

\subsection{Ablation Studies}

Ablations reuse the same HolisQA subset, preprocessing, generator, evaluator, and encoder. Each changes the named component; AR/CR aggregation excludes failed evaluator calls.

\begin{figure*}[!t]
\centering
\begin{minipage}[t]{0.49\textwidth}
\centering
\includegraphics[width=\linewidth]{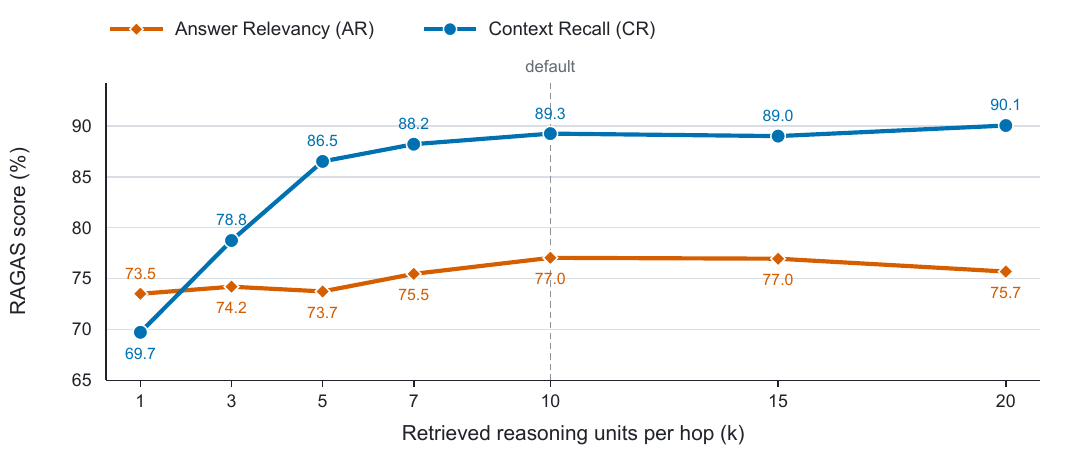}\\[-2pt]
\textbf{(a)} Per-hop candidate budget
\end{minipage}\hfill
\begin{minipage}[t]{0.49\textwidth}
\centering
\includegraphics[width=\linewidth]{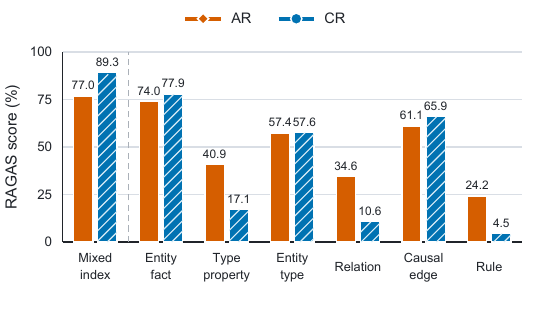}\\[-2pt]
\textbf{(b)} Reasoning-unit representation
\end{minipage}
\caption{Ablations on a fixed HolisQA subset. (a) CR improves rapidly with the number of retrieved units and then saturates, while AR peaks near $k=10$. (b) Restricting retrieval to any single unit kind underperforms the complete mixed index.}
\label{fig:ablation-summary}
\end{figure*}

\paragraph{Per-hop candidate budget.}
Figure~\ref{fig:ablation-summary}(a) varies $k$ while holding the query plan and all other settings fixed. CR rises from 69.7 at $k=1$ to 86.5 at $k=5$ and 89.3 at $k=10$. Increasing the budget further yields only small, non-monotonic changes: CR is 89.0 at $k=15$ and 90.1 at $k=20$. AR remains comparatively stable, reaching 77.0 at both $k=10$ and $k=15$, before declining to 75.7 at $k=20$. Thus, $k=10$ offers a practical balance between answer relevance, evidence coverage, and context size.

\paragraph{Typed unit diversity.}
Figure~\ref{fig:ablation-summary}(b) compares the complete mixed index with six variants that retain only one unit kind. The mixed index obtains 77.0 AR and 89.3 CR. Among the single-kind variants, \textsc{entity-fact} is strongest (74.0 AR and 77.9 CR), followed by \textsc{causal-edge} (61.1 AR and 65.9 CR). No individual representation matches the mixed index. The substantially lower scores of type properties, generic relations, and rules indicate that these representations alone provide insufficient evidence coverage. These results support treating the six unit kinds as complementary evidence representations rather than interchangeable schemas.

\begin{table}[!t]
\centering
\small
\setlength{\tabcolsep}{8pt}
\caption{Effect of hop-wise query decomposition on a fixed HolisQA subset. The full method uses $k$ candidates per hop; the single-query variant has one set of $k$.}
\label{tab:ablation-hop-decomposition}
\begin{tabular}{lrr}
\toprule
Variant & AR & CR \\
\midrule
Full \method & \textbf{77.03} & \textbf{89.30} \\
w/o hop decomposition & 70.04 & 42.33 \\
\bottomrule
\end{tabular}
\end{table}

\paragraph{Hop decomposition.}
Table~\ref{tab:ablation-hop-decomposition} replaces the ordered subquestions with a single retrieval over the original query. Removing decomposition reduces AR by 6.99 points and CR by 46.97 points. The larger CR loss supports decomposed evidence access, while the smaller AR loss suggests that an answer can remain topical despite missing support. Because the full method also allocates candidates across multiple hops, this experiment does not isolate query specificity from total evidence budget.

\subsection{Efficiency}

Unit extraction increases offline tokens; planning, hop solving, and composition add online latency. Appendix~\ref{sec:efficiency-details} gives the cost table, with CausalRAG2 efficiency left as unverified paper-reported values.

\section{Discussion and Future Work}

\paragraph{Long-context multi-chain reasoning.}
\method is architecturally suited to long contexts because its offline construction processes each passage independently. Moving from a short to a long corpus changes the number of indexed units, but not the amount of text handled by any single construction call. At inference, the complete corpus is never placed in the model context; only the top-$k$ units for at most $H$ information needs enter the reasoning buffer. The composer's context therefore scales with $O(Hk)$ rather than corpus length, allowing the same reasoning procedure to operate over short and long collections.

\paragraph{Efficiency.}
Efficiency is a part that we can study further. Future systems could batch or distill extraction and planning, cache reusable plans and embeddings, update only passages whose source spans change, and choose the hop count and per-hop candidate budget adaptively. These changes would preserve the bounded reasoning buffer while reducing both preprocessing tokens and online latency.

\paragraph{Scaling to very large corpora.}
Long-context tolerance at inference does not make complete corpus indexing free. For heterogeneous collections at Wikipedia scale, constructing a fine-grained index over every domain may be unnecessary. A promising extension is \emph{query-conditioned coarse indexing}: first route a query to a small set of plausible topical shards---for example, technology rather than politics or history---and then construct or activate the entity-structured index only within those shards. When routing is uncertain, cross-domain fallback retrieval can expand the search scope, preventing efficiency gains from making potentially relevant evidence irrecoverable.

\paragraph{Data sources.}
All datasets used in our experiments were obtained from publicly available benchmark releases and publicly accessible repositories.

\section{Conclusion}

We presented \method, a multi-chain reasoning framework that uses entity-structured index to find evidence and a compact buffer to reason over it. Focused hop retrieval narrows attention, source links keep every unit auditable, and delayed commitment prevents a locally attractive candidate from prematurely fixing the global answer. Across scientific and general-domain QA, the results and ablations show that this combination improves evidence coverage while making the reasoning process easier to inspect. More broadly, separating evidence addressing from final reasoning provides a practical foundation for extending multi-chain QA to longer and more heterogeneous corpora.

\section*{Limitations}

A limitation of this study is that we do not empirically evaluate \method on
long-context multi-chain reasoning. Existing long-context benchmarks primarily
test multi-document question answering or fan-out evidence aggregation
\citep{bai2024longbench,zhu2024fanoutqa,wang2024loong}; to our knowledge, they
do not isolate the setting in which multiple distinct evidence chains must be
preserved and reconciled within a single answer. We therefore lack an
appropriate benchmark for directly validating the advantage under long-context multi-chain conditions. Our MS~MARCO and QASC evaluations are also limited to 500 and 426 queries, respectively, rather than the complete evaluation splits. These subset results may not capture performance across the full range of questions; extending evaluation to the complete splits remains important when time and computational resources permit. In addition, the offline
index is most suitable when one corpus serves many questions: its construction
cost may not be well amortized for small or rapidly changing collections.
Query planning, hop solving, and final composition also require several staged
online calls. Finally, because \method is evaluated partly on scientific and
medical questions, its outputs should support rather than replace expert
judgment. Source grounding improves auditability but does not eliminate errors
caused by incomplete corpora or mistakes in extraction, planning, and
generation.

\bibliography{ensi_reason_refs}
\clearpage
\appendix
\begin{table*}[!t]
\section{Experimental Protocol}

This appendix provides the benchmark, baseline, efficiency, metric, and implementation details omitted from the main paper.

\subsection{Benchmark Details}\label{sec:benchmark-details}

\centering
\small
\setlength{\tabcolsep}{6pt}
\renewcommand{\arraystretch}{1.34}
\begin{tabularx}{\textwidth}{@{}p{0.13\textwidth}p{0.20\textwidth}p{0.29\textwidth}X@{}}
\toprule
\textbf{Benchmark} & \textbf{Domain and source} & \textbf{Question and reasoning focus} & \textbf{Evidence setting} \\
\midrule
HolisQA \citep{wang2026causalrag2} & Scientific articles spanning biology, business, computer science, medicine, and psychology & Open-ended holistic questions whose answers require integrating statements distributed across an article, including entity properties, relations, conditions, and causal processes & Gold answers and multiple supporting sentences are linked to the source articles. We evaluate the complete released dataset and report results separately for its five domains. \\
\hline
MS~MARCO \citep{bajaj2016msmarco} & Anonymized queries sampled from a production web-search system, paired with passages from retrieved web documents & Open-ended question answering in which a system must identify useful evidence among noisy, partially relevant search passages and produce a concise response & Candidate web passages accompany each query, and human annotators provide reference answers grounded in passages judged to contain sufficient information. \\
\hline
QASC \citep{khot2020qasc} & Grade-school science questions paired with a large corpus of elementary scientific facts & Multiple-choice questions are designed so that two complementary facts must be composed, commonly by connecting a concept in the question to the correct answer through an intermediate concept & Gold fact pairs expose the two-step evidence chain supporting each answer, allowing context recall to measure whether both complementary pieces of evidence were retrieved. \\
\bottomrule
\end{tabularx}
\caption{Benchmarks used in the main evaluation and the distinct evidence-reasoning capabilities they test.}
\label{tab:benchmark-details}
\end{table*}

\clearpage
\begin{table*}[!t]
\subsection{Baseline Details}\label{sec:baseline-details}

\centering
\small
\setlength{\tabcolsep}{6pt}
\renewcommand{\arraystretch}{1.32}
\begin{tabularx}{\textwidth}{@{}p{0.19\textwidth}p{0.18\textwidth}X@{}}
\toprule
\textbf{Baseline} & \textbf{Family} & \textbf{Retrieval and reasoning mechanism} \\
\midrule
NaiveGeneration \citep{wang2026causalrag2} & No retrieval & Generates an answer directly from the question, providing a retrieval-free reference for measuring the contribution of external evidence. \\
BM25 \citep{wang2026causalrag2} & Sparse retrieval & Ranks passages by lexical term matching and inverse document frequency, then supplies the highest-ranked passages to the answer generator. \\
StandardRAG \citep{lewis2020rag} & Dense retrieval & Encodes the question and passages in a shared semantic space, retrieves the nearest passages, and conditions generation on their text. \\
\hline
GraphRAG Global \citep{edge2024graphrag} & Graph-based RAG & Aggregates community reports across a corpus-level entity graph to answer broad questions requiring global thematic evidence. \\
GraphRAG Local \citep{edge2024graphrag} & Graph-based RAG & Starts from query-matched entities and retrieves their local graph neighborhoods, associated relations, claims, and source text. \\
LightRAG \citep{guo2024lightrag} & Graph-based RAG & Builds a lightweight entity--relation graph and combines local entity retrieval with global concept-level retrieval. \\
\hline
HippoRAG2 \citep{gutierrez2025hipporag2} & Structural RAG & Treats the corpus as a non-parametric associative memory and propagates activation over linked concepts for multi-step retrieval. \\
LeanRAG \citep{wang2026causalrag2} & Structural RAG & Organizes extracted knowledge into a lightweight hierarchical graph and retrieves evidence across semantically related graph regions. \\
CausalRAG \citep{wang2026causalrag2} & Causal RAG & Extracts causal relations and uses them to augment evidence retrieval, without explicit hierarchical causal reasoning. \\
CausalRAG2 \citep{wang2026causalrag2} & Causal RAG & Constructs hierarchical causal knowledge graphs and applies top-down causal pruning to select evidence for answer generation. \\
\bottomrule
\end{tabularx}
\caption{Baselines transcribed from the CausalRAG2 evaluation suite. Their reported scores, efficiency values, and costs are taken from that work.}
\label{tab:baseline-details}
\end{table*}

\clearpage
\begin{table*}[!t]
\subsection{Efficiency Details}\label{sec:efficiency-details}

\centering
\footnotesize
\setlength{\tabcolsep}{7pt}
\renewcommand{\arraystretch}{1.18}
\caption{Cost and latency on HolisQA. Baseline entries are transcribed from the CausalRAG2 efficiency table and were not independently reproduced; only the \method row is measured in our environment. Construction values are normalized per 1K corpus tokens, online values per query, and evaluation cost is excluded.}
\label{tab:efficiency}
\begin{tabular*}{\textwidth}{@{\extracolsep{\fill}}lrrr|rrr@{}}
\toprule
& \multicolumn{3}{c|}{Construction (per 1K corpus tokens)} & \multicolumn{3}{c}{Online query (per query)} \\
Method & Tokens & Time (s) & Cost (\$) & Tokens & Time (s) & Cost (\$) \\
\midrule
NaiveGeneration & 0.00 & 0.000 & 0.00000 & 553.63 & 0.354 & 0.000078 \\
BM25            & 0.00 & 0.000 & 0.00000 & 967.90 & 0.562 & 0.000110 \\
StandardRAG     & 0.00 & 0.000 & 0.00000 & 1016.56 & 0.638 & 0.000120 \\
GraphRAG Global & 20241.49 & 19.694 & 0.00340 & 6566.85 & 1.850 & 0.000680 \\
GraphRAG Local  & 20241.49 & 19.694 & 0.00340 & 4968.42 & 1.306 & 0.000510 \\
LightRAG        & 11245.55 & 7.317 & 0.00150 & 1517.70 & 0.487 & 0.000180 \\
HippoRAG2       & 5404.08 & 4.647 & 0.00092 & 3125.54 & 0.641 & 0.000320 \\
LeanRAG         & 8893.11 & 5.589 & 0.00120 & 5031.30 & 0.731 & 0.000460 \\
CausalRAG       & 9764.49 & 9.022 & 0.00140 & 5468.44 & 0.849 & 0.000520 \\
CausalRAG2      & 9864.49 & 9.622 & 0.00150 & 5075.65 & 0.801 & 0.000490 \\
\midrule
\method         & 14328.85 & 6.880 & 0.00261 & 30333.35 & 2.601 & 0.00214 \\
\bottomrule
\end{tabular*}

The offline token increase reflects two LLM stages that organize passages and extract source-grounded, typed units. Online cost is dominated by planning, per-hop solving, and final composition rather than dense retrieval. The values should not be read as a hardware-controlled head-to-head comparison: the baseline rows are paper-reported, and the released CausalRAG2 pipeline did not complete a full-scale local run in our environment.
\end{table*}

\clearpage
\subsection{RAGAS Evaluation Metrics}\label{sec:ragas-metrics}

We use RAGAS answer relevancy (AR) and context recall (CR) \citep{es2024ragas}. Both metrics are computed on a $0$--$1$ scale and reported as percentages in the main tables.

\paragraph{Answer relevancy.}
AR measures whether the generated answer directly addresses the input question. RAGAS generates $N$ synthetic questions $g_i$ from the answer and compares their embeddings with that of the original question $q$:
\begin{equation}
\operatorname{AR}=\frac{1}{N}\sum_{i=1}^{N}\cos\!\left(E(g_i),E(q)\right).
\end{equation}
We use $N=3$, corresponding to strictness~3. A high score indicates that the response is focused and pertinent; incomplete, tangential, or unnecessarily verbose responses receive lower scores. AR does not by itself establish factual correctness: an incorrect answer can still be relevant if it directly addresses the question.

\paragraph{Context recall.}
CR measures whether retrieval recovers the information required by the reference answer. The evaluator decomposes the reference answer into claims and judges whether each claim is attributable to the retrieved source passages:
\begin{equation}
\operatorname{CR}=\frac{N_{\mathrm{supported}}}{N_{\mathrm{reference}}}.
\end{equation}
Here, $N_{\mathrm{supported}}$ is the number of reference claims attributable to retrieved context, and $N_{\mathrm{reference}}$ is the total number of reference claims. A high CR therefore means that little required evidence was missed. It evaluates evidence coverage rather than response quality and does not penalize irrelevant retrieved passages. AR and CR are complementary: CR tests whether the system retrieved sufficient support, while AR tests whether the generated response remains aligned with the question.

\clearpage
\subsection{Implementation Details}

\paragraph{Offline construction.}
Each source article is segmented into sentence-aligned units and passed to an LLM passage constructor that returns contiguous entity-centered spans. The constructor records the main entity, entity type, semantic focus, evidence role, and source sentence IDs. These output records carry an explicit semantic-passage marker, and the reasoning runner preserves every marked record as one extraction input. Its fixed-character chunker is invoked only for unmarked compatibility inputs and is not used for the reported semantic-passage runs. A second LLM pass extracts typed reasoning units from each passage. The normalizer assigns a stable unit ID, canonical predicate, aliases, scope, confidence, evidence quote, and provenance pointer. The passage and extraction prompts never receive question, answer, or gold-context fields.

\paragraph{Online reasoning.}
The query planner returns the reasoning mode, expected answer type, entity mentions, properties, constraints, and an ordered list of one-hop subquestions. For each hop, the retriever embeds the concatenation of the subquestion, target, and bridge terms, compares it with the cached embeddings of all reasoning units, and retains the top-$k$ candidates. The main configuration uses $k=10$; the budget ablation changes only this value. Each hop solver sees only its own subquestion, planner metadata, and candidate set; it does not receive earlier hop answers or gold evidence. Retrieved units are mapped back to their source passages, packed within the evidence budget, and selected for final composition. The final composer receives only the original question and selected source passages; local answers remain in the audit trace but are not passed to the composer. No separate verifier is invoked in this backend.

\paragraph{Models and reproducibility.}
The dense encoder is \texttt{all-MiniLM-L6-v2} \citep{reimers2019sbert}. Reported \method runs explicitly select the hop-wise candidate backend, which skips the codebase's legacy graph-construction path. Generation stages use GPT-5-nano, and evaluation uses RAGAS~0.4.3 with Gemini-3.5-Flash-Lite. We serialize passages, units, auxiliary indexes, query plans, embedding caches, per-hop candidates, local answers, final answers, retrieved contexts, and evaluator inputs so that each result can be traced to its model and evidence state. The published CausalRAG2 rows use the evaluator available to that work and are not presented as a same-runtime rerun.

We initially attempted to use Gemini-2.5-Flash-Lite, but the API reported
that it was no longer available to new users, preventing access through
our account; we therefore used Gemini-3.5-Flash-Lite. In a local rerun
on a Biology subset of HolisQA, the resulting scores were close to those
reported by CausalRAG2, providing a sanity check for our evaluation
pipeline and supporting the use of its published results as reference
comparisons, without establishing identical evaluation conditions.

\paragraph{Metrics.}
AR and CR use the RAGAS implementation and its configured judge model. HolisQA additionally reports token-level answer-overlap F1 against reference answers; it complements the judge-based metrics but is sensitive to wording. Sentence-ID recall is retained only as a provenance diagnostic rather than substituted for RAGAS context recall. Latency is measured from the first online model call through final composition, excluding one-time offline preprocessing unless explicitly stated.

\clearpage
\subsection{Algorithms}\label{sec:algorithms}

The following pseudocode summarizes the executed candidate-set backend. It is intentionally expressed in terms of serialized passages, reasoning units, and ordered hop records rather than graph nodes or edges. In Algorithm~\ref{alg:retrieval}, \textsc{RetrievalText} concatenates a subquestion with its target and bridge terms.

\begin{algorithm}[H]
\footnotesize
\caption{Offline Reasoning-Unit Construction}
\label{alg:offline}
\begin{algorithmic}[1]
\Require Corpus $D$; passage constructor $\mathcal{P}$
\Require Unit extractor $\mathcal{X}$; normalizer $\mathcal{N}$; encoder $E$
\Ensure Unit store $U$; indexes $I$; embedding matrix $M$
\State $U \gets \emptyset$; $I \gets \emptyset$
\ForAll{documents $d \in D$}
    \State $S \gets \Call{SentenceUnits}{d}$
    \State $P_d \gets \mathcal{P}(S)$
    \ForAll{passages $p \in P_d$}
        \State $R \gets \mathcal{X}(p)$
        \ForAll{raw units $r \in R$}
            \State $u \gets \mathcal{N}(r, \Call{Provenance}{p})$
            \State assign a stable unit identifier to $u$
            \State $U \gets U \cup \{u\}$
            \State $I \gets \Call{UpdateTypedIndexes}{I,u}$
        \EndFor
    \EndFor
\EndFor
\State $M \gets E(\{\Call{EmbeddingText}{u}:u\in U\})$
\State \Call{Serialize}{$U,I,M$}
\State \Return $U,I,M$
\end{algorithmic}
\end{algorithm}

\newpage

\begin{algorithm}[H]
\footnotesize
\caption{Hop-Wise Candidate Retrieval}
\label{alg:retrieval}
\begin{algorithmic}[1]
\Require Question $q$; planner $\mathcal{L}$; unit store $U$
\Require Embeddings $M$; encoder $E$
\Require Budget $k$; maximum hops $H$
\Ensure Query plan $P$; ordered candidate trace $T$
\State $P \gets \mathcal{L}(q,H)$; $T \gets \emptyset$
\ForAll{$s_h \in P.\textsc{Subquestions}$ in order}
    \State $x_h \gets \Call{RetrievalText}{s_h}$
    \State $v_h \gets E(x_h)$
    \State $z_h \gets Mv_h$
    \State $C_h \gets \Call{TopK}{U,z_h,k}$
    \State append $(s_h,C_h,z_h[C_h])$ to $T$
\EndFor
\State \Return $P,T$
\end{algorithmic}
\end{algorithm}

\begin{algorithm}[H]
\footnotesize
\caption{Candidate-Set Answer Generation}
\label{alg:generation}
\begin{algorithmic}[1]
\Require Question $q$; query plan $P$; candidate trace $T$; provenance $\pi$
\Require Hop solver $\mathcal{L}_{\mathrm{hop}}$; source selector $\mathcal{S}$; composer $\mathcal{L}_{\mathrm{final}}$
\Ensure Final answer $a$; auditable intermediate trace $A$
\State $A \gets \emptyset$
\ForAll{$(s_h,C_h) \in T$}
    \State $a_h \gets \mathcal{L}_{\mathrm{hop}}(s_h,C_h)$
    \State append $(s_h,C_h,a_h)$ to $A$ \Comment{Audit only}
\EndFor
\State $R \gets \Call{PackSourcePassages}{T,\pi,16000}$
\State $J \gets \mathcal{S}(q,\Call{TitlesAndSnippets}{R})$
\If{$J$ contains no valid article identifiers}
    \State $J \gets \Call{TopTermMatchedGroups}{q,R,3}$
\EndIf
\State $B \gets \Call{PassagesOfGroups}{R,J}$
\State $a \gets \mathcal{L}_{\mathrm{final}}(q,B)$
\State \Call{Serialize}{$P,T,A,B,a$}
\State \Return $a,A$
\end{algorithmic}
\end{algorithm}

\clearpage
\subsection{Prompt Templates}\label{sec:prompts}
\begingroup
\sloppy

The boxes below reproduce the principal system prompts used by the released implementation. Runtime values such as the article text, question, query plan, and retrieved candidates replace the indicated placeholders. Gold answers and gold evidence are never included in construction, planning, retrieval, or generation prompts.

\prompttitle{Passage Construction: System Prompt}
\begin{lstlisting}[style=promptcode]
You are the offline passage-construction module of RESIN.

Construct query-independent passages from a scientific article. Do not answer
any question and do not summarize the whole article.

RESIN passage contract:
- Each passage has one primary entity, mechanism, intervention, finding,
  method, outcome, or other evidence-bearing topic.
- Keep adjacent sentences together when they jointly define the entity or
  explain its properties, relations, mechanism, or result.
- Split when the main evidence center changes.
- Passage lengths are semantic and variable. Do not use a fixed number of
  sentences, tokens, or characters as the passage definition.
- Return only contiguous ranges of the supplied source sentences.
- Do not paraphrase, reorder, or add outside knowledge.
- Unit IDs are 1-based integers. Copy start_unit and end_unit from the
  provided unit labels. Valid IDs range from 1 to N.

Return JSON only in this form:
{"passages":[{"start_unit":1,"end_unit":3,"main_entity":"...",
"entity_type":"...","semantic_focus":"...","evidence_role":"..."}]}
\end{lstlisting}

\prompttitle{Reasoning-Unit Extraction: System Prompt}
\begin{lstlisting}[style=promptcode]
You are the offline extractor for RESIN.
Return JSON only: {"units":[...]}.

RESIN represents multi-chain reasoning as typed states and directed
transitions. Extract the smallest evidence-backed units that can be composed.
Each unit must have:
kind: entity_fact, type_property, entity_type, causal_edge, relation, or rule;
subject; subject_kind: entity, type, variable, event, state, or unknown;
predicate; object; conditions; conclusion; source_state; target_state;
aliases; scope; modality; polarity; confidence; evidence_quote.

Use these conventions:
- entity_fact: an ordinary fact about a specific entity.
- type_property: a generic property of a type, such as Humans are mortal.
- entity_type: an explicit membership assertion, such as Socrates is a human.
- causal_edge: a directed transition, such as enzyme deficiency causes substrate
  accumulation. Use source_state and target_state when the sentence expresses
  property-value states. Preserve the direction and any modality such as may,
  can, or is thought to.
- rule: only a conditional statement with non-empty conditions and a conclusion,
  such as If people drink more water, health improves. Do not use rule for a
  simple one-edge causal fact.

For a sentence containing multiple causal steps, emit multiple causal_edge
units, one per supported transition, while keeping the same evidence_quote.
Do not infer a transition merely because two entities occur in the same text.
Do not merge a statue with the person it represents. Keep entity type,
context, polarity, and uncertainty explicit. Every unit needs an exact short
evidence_quote.
\end{lstlisting}

The extraction user message contains only article metadata and the delimited article chunk: \texttt{ARTICLE\_METADATA: \{metadata\}} and \texttt{ARTICLE\_CHUNK: <<<\{text\}>>>}.

\clearpage
\prompttitle{Query Planning: System Prompt}
\begin{lstlisting}[style=promptcode]
You are the RESIN query planner. Do not answer the question.
Decompose the question into an ordered list of minimal one-hop subquestions.
Each subquestion must be answerable from a small set of evidence-backed
reasoning units and should expose the bridge needed by the next hop. Do not
write a multi-hop subquestion. Use at most the requested MAX_HOPS.
Return JSON only:
{"reasoning_mode":"entity_join | type_inheritance | conditional_rule |
causal_chain | temporal | comparison | mixed",
 "answer_type":"entity | boolean | date | explanation | unknown",
 "target_description":"","entity_mentions":[],"type_hints":[],
 "properties":[],"constraints":[],"source_constraints":[],
 "time_constraints":[],"bridge_terms":[],
 "subquestions":[{"hop":1,"question":"","target":"",
 "depends_on":[],"bridge_terms":[],"reasoning_type":""}]}
\end{lstlisting}

The corresponding user template is \texttt{MAX\_HOPS: \{max\_hops\}} followed by \texttt{QUESTION: \{query\}}.

\prompttitle{Hop Solving: System Prompt}
\begin{lstlisting}[style=promptcode]
You are a RESIN hop solver.
Answer only the supplied one-hop subquestion using the candidate reasoning
units for this hop. Do not use candidates from another hop and do not add
outside knowledge. The answer should expose the entity, property, value, or
bridge needed by later hops. If the candidates do not support the subquestion,
return insufficient rather than guessing.
Return JSON only:
{"answer":"","status":"supported | insufficient | conflicted",
 "supporting_unit_ids":[],"reasoning_step":"","confidence":0.0}
\end{lstlisting}

At runtime, the user payload contains the hop number, subquestion, target, dependency metadata, and that hop's ranked candidate reasoning units. It contains neither gold evidence nor gold answers.

\clearpage
\prompttitle{Final Composition: System Prompt}
\begin{lstlisting}[style=promptcode]
Use only the retrieved source passages to answer the research question.
Return JSON with answer first: {"answer":"...","status":"supported | insufficient","supporting_unit_ids":[]}.
Form a self-contained answer of two to four short sentences, about 55-85 words. Cover each requested factor, mechanism, comparison, outcome, and condition found in the passages. Reuse the source's exact technical terms and short phrases for the central findings; avoid broad synonyms that erase the findings. Preserve polarity and uncertainty.
Begin with the substantive answer. Do not describe the analysis, give background definitions, list unrelated possibilities, cite sources, or add a generic conclusion. If evidence is partial, give the supported facts without a long caveat. Never use a reference answer or outside knowledge.
\end{lstlisting}

The final user payload contains only the original question and selected source passages. It does not contain the query plan, candidate-unit records, or generated hop answers.
\endgroup

\clearpage
\begingroup
\makeatletter
\setlength{\@fptop}{0pt}
\makeatother
\begin{table*}[p]
\centering
\normalsize
\setlength{\tabcolsep}{6pt}
\renewcommand{\arraystretch}{1.15}
\section{Case Studies}
\label{sec:diagnostic-cases}
\noindent We present selected cases in which CausalRAG2 omits a required evidence component and \method recovers it. The comparison examines original source evidence, saved CausalRAG2 prompts, and the latest \method answers. Each table isolates the missing component and its architectural significance; it does not equate recovering that component with complete correctness of every additional statement.
\par\medskip
\caption{Kandelia: preserving molecular findings alongside the dominant physiological explanation.}
\label{tab:case-kandelia}
\begin{tabularx}{\textwidth}{@{}>{\raggedright\arraybackslash}p{0.19\textwidth}>{\raggedright\arraybackslash}X@{}}
\toprule
\textbf{Case component} & \textbf{Details} \\
\midrule
\textbf{Question} & What mechanisms explain low-temperature tolerance in the cold-tolerant \emph{Kandelia obovata} population, and what molecular insights were obtained? \\
\midrule
\textbf{Required evidence} & The answer needs antioxidant/metabolic adaptation, transcription-factor findings, and SNP/alternative-splicing evidence. A physiological explanation alone does not cover the requested molecular insights. \\
\midrule
\textbf{CausalRAG2 evidence} & The exported recall context includes SNP and splicing information. The final generation prompt instead retains antioxidant activity, named transcription factors, and metabolic/signaling pathways, without the SNP/splicing findings. \\
\midrule
\textbf{CausalRAG2 response} & The answer explains catalase-driven ROS management, WRKY/NAC/MYB/ERF expression, and enriched pathways, but does not report SNP or alternative-splicing findings. \textcolor{red!70!black}{\textbf{Missing molecular evidence.}} \\
\midrule
\textbf{Why the omission occurs} & The recorded evidence assembly removes a distinct molecular evidence category before final generation. The answer then elaborates the retained antioxidant/transcriptional explanation rather than recovering the missing category. The trace locates this loss between broad recall and the final prompt, not at initial extraction. \\
\midrule
\textbf{RESIN organization} & Typed units make molecular findings independently addressable. Decomposed needs can distinguish the physiological mechanism from molecular evidence; the latter need not depend on selection of the former's causal chain. \\
\midrule
\textbf{RESIN source evidence} & The latest mapped passages preserve both the transcription-factor findings and the statement that SNP diversity and alternative splicing may enhance protein function and contribute to cold tolerance. \\
\midrule
\textbf{RESIN response} & The answer retains antioxidant/metabolic responses and the named transcription factors, and additionally reports potential SNP/splicing contributions to cold tolerance. \textcolor{green!40!black}{\textbf{Missing evidence category recovered.}} It does not further explain the proposed protein-function contribution. \\
\midrule
\textbf{Reference answer} & The reference includes antioxidant and metabolic regulation, transcriptional/post-transcriptional responses, named TF families, and possible SNP/splicing contributions through enhanced protein function. \\
\midrule
\textbf{Key difference} & Relevant evidence spans several molecular roles rather than one dominant mechanism. \method provides separate access to these roles and recovers their original passages for composition. The latest answer includes a category that is lost from CausalRAG2's final evidence set. \\
\bottomrule
\end{tabularx}
\end{table*}

\begin{table*}[p]
\centering
\normalsize
\setlength{\tabcolsep}{6pt}
\renewcommand{\arraystretch}{1.15}
\caption{CD1c: retaining a comparative structural property during answer composition.}
\label{tab:case-cd1c}
\begin{tabularx}{\textwidth}{@{}>{\raggedright\arraybackslash}p{0.19\textwidth}>{\raggedright\arraybackslash}X@{}}
\toprule
\textbf{Case component} & \textbf{Details} \\
\midrule
\textbf{Question} & How does CD1c present lipids to TCRs, and how does this mechanism differ from other antigen-presenting molecules? \\
\midrule
\textbf{Required evidence} & The mechanism combines simultaneous two-lipid loading with sideways presentation of bulky headgroups through a side portal. Two-lipid loading is a distinguishing property, not merely another step in the presentation pathway. \\
\midrule
\textbf{CausalRAG2 evidence} & Its final generation prompt contains a fragment stating that two lipid antigens are presented from the top and side of the cleft. Its reranking draft also mentions two lipids. The required cardinality fact is therefore available before the final answer. \\
\midrule
\textbf{CausalRAG2 response} & The answer describes lipid binding, sideways headgroups, the side portal, and TCR recognition, but omits simultaneous two-lipid loading. \textcolor{red!70!black}{\textbf{Missing distinguishing property.}} \\
\midrule
\textbf{Why the omission occurs} & Available evidence is not fully carried into final synthesis: the answer explains how presentation works without preserving a central property needed to explain how it differs. This is a composition omission, not a failure to retrieve the two-lipid fact. \\
\midrule
\textbf{RESIN organization} & An entity-specific property can be represented separately from the mechanistic transition. Need-specific retrieval locates the explanatory and comparative claims; source selection recovers original passages for direct final composition, without passing generated hop answers. \\
\midrule
\textbf{RESIN source evidence} & The latest mapped passage explicitly states that two lipids, rather than one, bind in the CD1c cleft, and that bulky headgroups protrude laterally through a side portal while permitting TCR recognition. \\
\midrule
\textbf{RESIN response} & The answer states that CD1c can bind two lipids simultaneously and explains sideways protrusion of bulky headgroups through the side portal. \textcolor{green!40!black}{\textbf{Distinguishing property recovered.}} Its additional peptide-MHC comparison is separate from this source-grounded finding. \\
\midrule
\textbf{Reference answer} & The reference combines two-lipid binding, sideways orientation, headgroup protrusion, and TCR recognition to explain the distinctive presentation mechanism. \\
\midrule
\textbf{Key difference} & Recovering a pathway does not automatically preserve the property that makes it distinctive. \method's interface gives explanatory and comparative evidence separate roles and keeps them jointly available; the latest answer preserves both in this case. \\
\bottomrule
\end{tabularx}
\end{table*}

\begin{table*}[p]
\centering
\normalsize
\setlength{\tabcolsep}{6pt}
\renewcommand{\arraystretch}{1.15}
\caption{CKL12: separating a growth phenotype from evidence of BR regulation.}
\label{tab:case-ckl12}
\begin{tabularx}{\textwidth}{@{}>{\raggedright\arraybackslash}p{0.19\textwidth}>{\raggedright\arraybackslash}X@{}}
\toprule
\textbf{Case component} & \textbf{Details} \\
\midrule
\textbf{Question} & How does CKL12 contribute to plant growth, and what evidence suggests regulation by brassinosteroid (BR) signaling? \\
\midrule
\textbf{Required evidence} & The answer requires a growth phenotype and evidence of regulation. Insertion mutants suggest a contribution to hypocotyl and primary-root growth; promoter binding and reduced transcript after transcription-factor knock-down support the BR connection. \\
\midrule
\textbf{CausalRAG2 evidence} & Its final prompt contains a BR-regulation fragment, including promoter sites and knock-down results, but no hypocotyl or primary-root finding. The saved trace does not identify which earlier processing step first excluded the phenotype. \\
\midrule
\textbf{CausalRAG2 response} & The answer says CKL12's exact growth role is unspecified in the provided items, then describes BR regulation. \textcolor{red!70!black}{\textbf{Growth component unanswered.}} \\
\midrule
\textbf{Why the omission occurs} & Regulation and phenotype are different evidence roles. Retaining the regulatory fragment does not supply the missing growth observation, so the final answer cannot connect the two requested parts using its provided evidence. \\
\midrule
\textbf{RESIN organization} & Growth effects and regulatory support are directly searchable claims, not interchangeable descriptions of CKL12. Need-specific retrieval allocates candidate capacity to each role; the final buffer can preserve the source passages supporting both roles. \\
\midrule
\textbf{RESIN source evidence} & The latest mapped passage contains the tentative hypocotyl/root finding, transcript-end observations, promoter binding sites, and the knock-down experiment. The phenotype is therefore represented in the source evidence associated with the latest run. \\
\midrule
\textbf{RESIN response} & The answer names hypocotyl and primary-root growth and includes the BR-related promoter sites. \textcolor{green!40!black}{\textbf{Missing phenotype recovered.}} It still omits insertion/knock-down details and strengthens the source's tentative growth wording. \\
\midrule
\textbf{Reference answer} & The reference says CKL12 appears to contribute to growth; insertion findings suggest a role for the transcript's $3'$ end, while promoter and knock-down results support downstream BR regulation. \\
\midrule
\textbf{Key difference} & A regulator-centered evidence set can answer the signaling part while missing the phenotype. \method treats the two needs separately and retains their claims together. This case illustrates recovery of the missing growth component, not complete recovery of every supporting experiment. \\
\bottomrule
\end{tabularx}
\end{table*}

\clearpage
\endgroup

\end{document}